\documentclass{article}

\usepackage[english]{babel}

\usepackage[letterpaper,top=2cm,bottom=2cm,left=3cm,right=3cm,marginparwidth=1.75cm]{geometry}

\usepackage{amsmath}
\usepackage{graphicx}
\usepackage{subfig}

\usepackage[colorlinks=true, allcolors=blue]{hyperref}

\title{ChemVise: Maximizing Out-of-Distribution Chemical Detection with the Novel Application of Zero-Shot Learning}
\author{Alexander M. Moore
\and Randy C. Paffenroth
\and Ken T. Ngo
\and Joshua R. Uzarski
}

\begin{document}
\maketitle

\begin{abstract}
Accurate chemical sensors are vital in medical, military, and home safety applications. Training machine learning models to be accurate on real world chemical sensor data requires performing many diverse, costly experiments in controlled laboratory settings to create a data set. In practice even expensive, large data sets may be insufficient for generalization of a trained model to a real-world testing distribution. Rather than perform greater numbers of experiments requiring exhaustive mixtures of chemical analytes, this research proposes learning approximations of complex exposures from training sets of simple ones by using single-analyte exposure signals as building blocks of a multiple-analyte space. We demonstrate this approach to synthetic sensor responses surprisingly improves the detection of out-of-distribution obscured chemical analytes. Further, we pair these synthetic signals to targets in an information-dense representation space utilizing a large corpus of chemistry knowledge. Through utilization of a semantically meaningful analyte representation spaces along with synthetic targets we achieve rapid analyte classification in the presence of obscurants without corresponding obscured-analyte training data. Transfer learning for supervised learning with molecular representations makes assumptions about the input data. Instead, we borrow from the natural language and natural image processing literature for a novel approach to chemical sensor signal classification using molecular semantics for arbitrary chemical sensor hardware designs.\end{abstract}

\section{Introduction}

\begin{figure}
    \centering
    \includegraphics[width=0.90\linewidth]{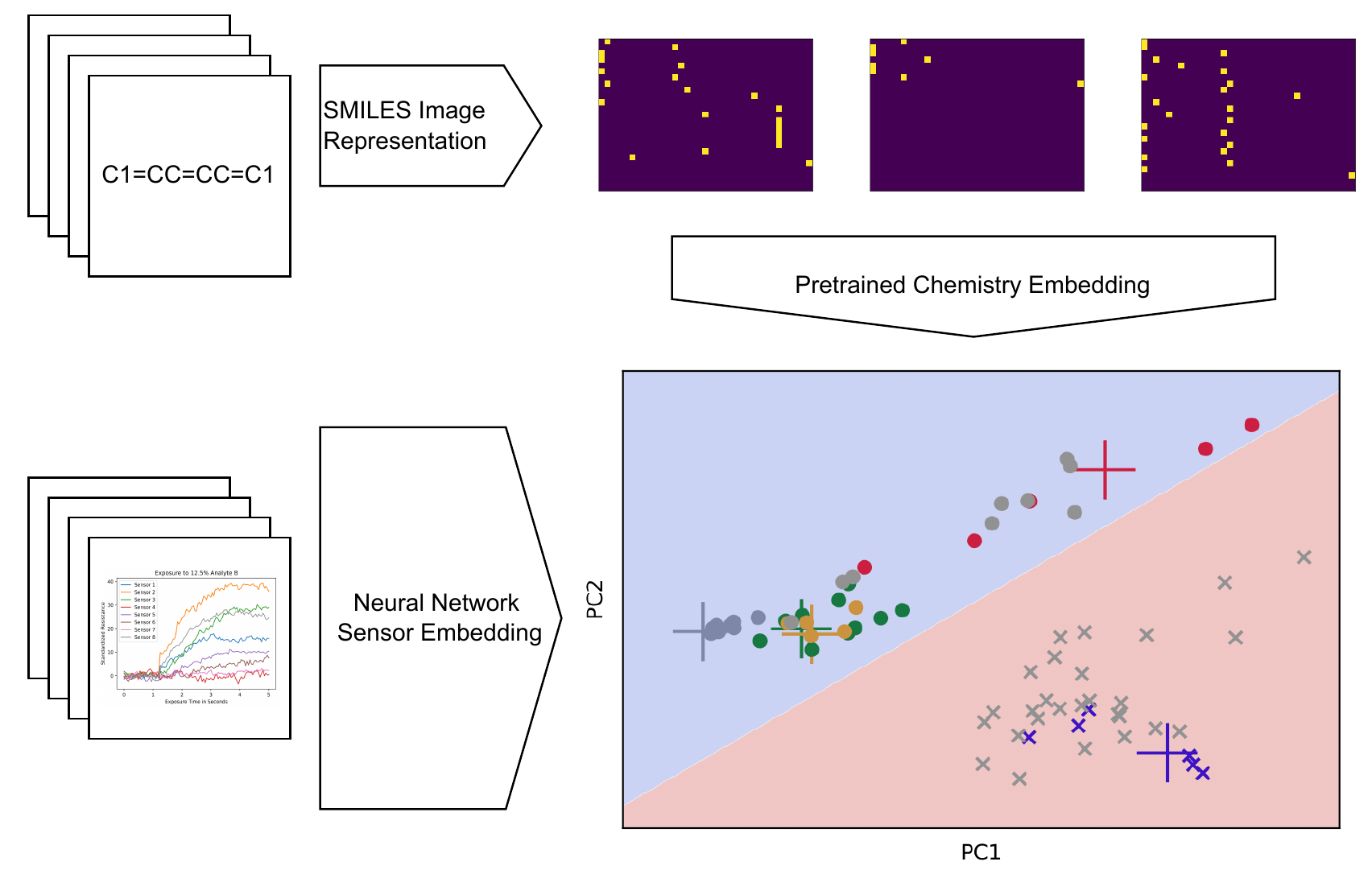}
    \caption{ChemVise utilizes activations from a pretrained supervised molecular property model to represent analytes of interest with geometric interpretation. The utilization of arithmetic means between points in analyte space creates improved, tunable decision boundaries for chemical sensor classifiers. We find that semantic chemistry representations outperforms non-chemical embeddings and leads to greater discriminability of obscured chemical analytes.}
    \label{fig:chemvise_embeddings}
\end{figure}

Classifying chemical vapors is vital to military, industrial, and safety applications. Hazardous chemical detection becomes challenging in the presence of obscurants and environmental factors as sensor responses change. This paper addresses the generalizability of data mining applied to challenging chemical vapor mixtures with novel deep learning approaches. We take inspiration from embedding-translation models in calling our approach ``ChemVise'', a portmanteau of the constituent Chemception \cite{goh2017} molecular attribute prediction model and DeViSE \cite{frome2013} visual-semantic embeddings. Rather than natural image and sentence semantics, we utilize deep representations of SMILES molecular images paired with chemiresistive sensor signals to learn improved representations for downstream predictors (Figure \ref{fig:chemvise_embeddings}). Improved chemical representations for chemical sensing allows even challenging double-analyte samples to be encoded such that they are linearly separable with simple classifiers. We find our approach significantly outperforms competitive models for the detection of obscured chemical analytes with only unobscured analytes as training data.

In the domain of chemical sensing, gathering data sufficient for complex chemical detection tasks is highly laborious and expensive \cite{wiederoder2017graphene, nallon2016discrimination}. The ability to import significant domain knowledge in the form of transfer learning is vital to the rapid development of diverse and accurate sensors. Promising results in chemistry-applied deep learning indicate that well-tuned models shorten the development cycle of new tools and in some cases can replace repetitive laboratory experiments \cite{wei2019}.

\begin{figure*}
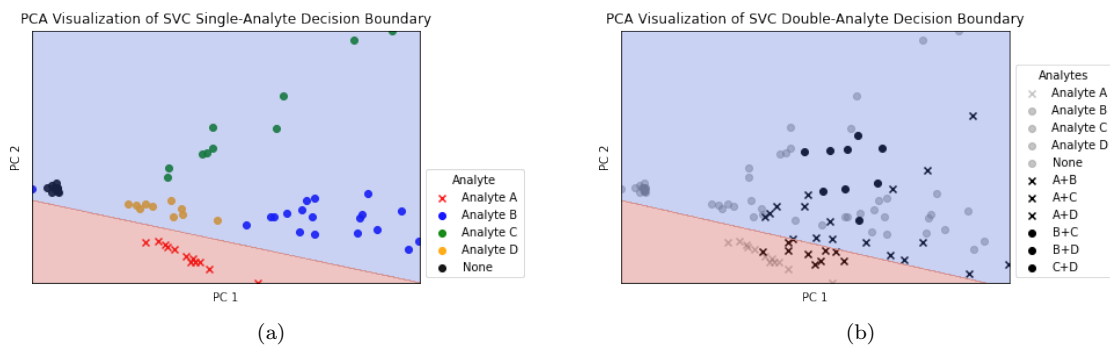

    \centering
    \subfloat[]{\label{fig:pca_svc_sub1}{\includegraphics[width=0.45\linewidth]{PCA_svc_decision_singles.pdf} }}%
    \qquad
    \subfloat[]{\label{fig:pca_svc_sub2}{\includegraphics[width=0.45\linewidth]{PCA_svc_decision_doubles.pdf} }}%
    \caption{Visualization of the shortcomings of linear decomposition with linear classifiers for mixture data. \textbf{Left}: Under a 2-dimensional PCA transformation, signal data of single-analyte exposures may be well-separated by a linear classifier. \textbf{Right}: Under a PCA embedding of signal data, the same model fails to accurately discriminate between samples containing Analyte A. Samples marked with x contain Analyte A and are intermixed with non-Analyte A single-analyte exposures, leading to poor classification results from this linear classifier.}%
    \label{fig:pca_svc}%
\end{figure*}

For our chemiresistive sensor signal classification task, an example machine learning approach to the discrimination of chemical Analyte A from chemical Analytes B, C, and D succeeds when one analyte is present at a time (Figure \ref{fig:pca_svc_sub1}). For the more complex detection test of obscured signals given by analyte mixtures, this approach performs poorly (Figure \ref{fig:pca_svc_sub2}). The ability to discriminate Analyte A signals from other analyte exposures even when obscured by a secondary analyte is a vital task for chemical sensors, and one which demands changes to standard data mining practice.

In order to classify more challenging analyte mixtures which do not appear in the training set, we borrow from zero-shot, transfer, and multi-modal learning approaches in deep natural image and language learning \cite{larochelle2008, frome2013, radford2021}. We contribute the following to the challenging task of obscured analyte detection for one chemical detection paradigm, with generalizations available for any sensing hardware design:

\begin{enumerate}
    \item We introduce the novel ChemVise approach which modifies transfer learning to utilize chemistry domain knowledge for any chemiresistive sensor data (Section \ref{sec:method}).
    
    \item We apply linear interpolation between single analyte exposures for analyte mixture representations to improve the detection of obscured analytes (Section \ref{sec:lc}). The utilization of linear interpolation for chemistry representations as well as sensor response approximations to mixture data is yet unexplored in the literature.
    
    \item We demonstrate that transfer learning of ChemVise outperforms other machine and deep learning approaches to chemical sensing as well as alternative non molecular-semantic representations (Section \ref{sec:results}).

    \item We provide an outline for the generalization of this approach to arbitrary sensing hardware. Whereas most transfer learning techniques would be domain specific, this approach can adapt the training of the deep encoder for arbitrary input data to the same latent representation space (Section \ref{sec:conclusion}).
\end{enumerate}

\begin{figure}
    \centering
    \includegraphics[width = 0.8\linewidth]{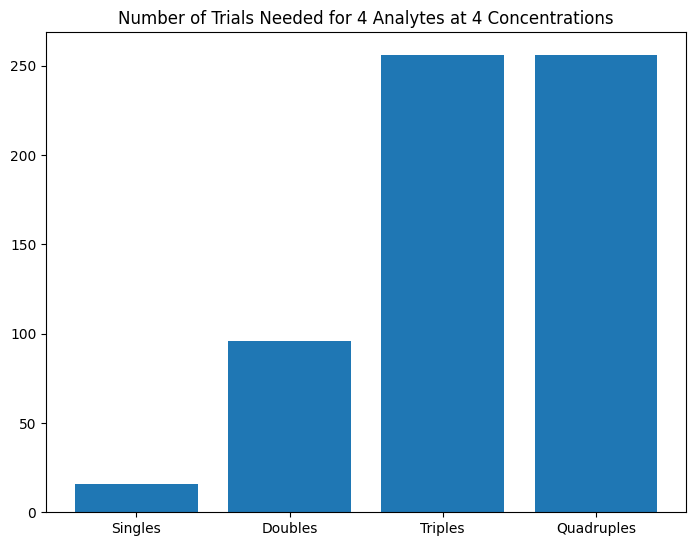}
    \caption{The corresponding number of experimental trials to be performed given four analytes (A, B, C, D) at four possible exposure magnitudes (12.5\%, 15\%, 20\%, 25\%). }
    \label{fig:num_experiments}
\end{figure}

\begin{figure*}
    \centering
    \subfloat{{\includegraphics[width=0.45\linewidth]{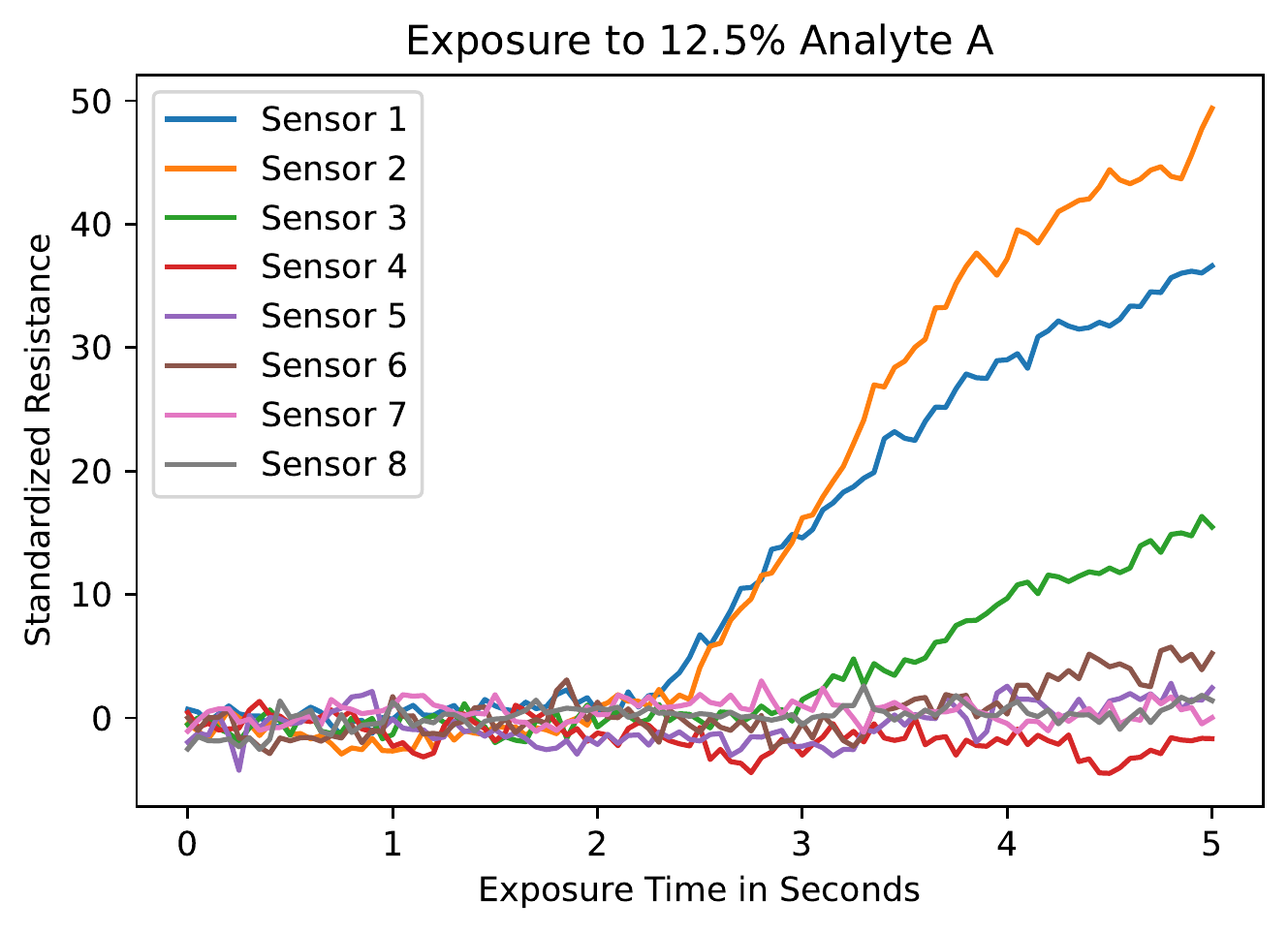} }}%
    \qquad
    \subfloat{{\includegraphics[width=0.45\linewidth]{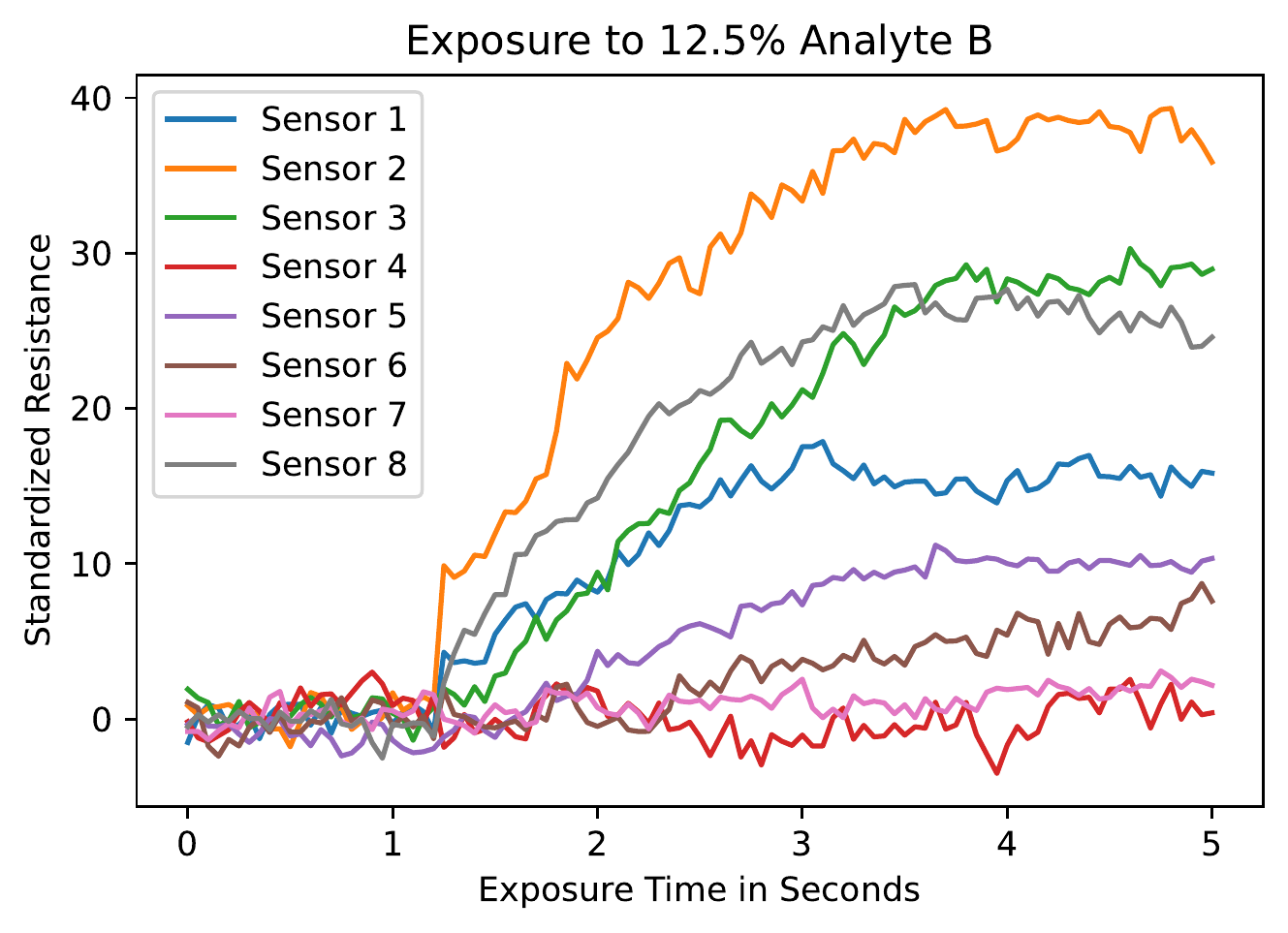} }}%
    \caption{Two example single-analyte exposures. \textbf{Left}: A 5-second window including the moment a mixture of 12.5\% Analyte A is exposed to the array of eight sensors. \textbf{Right}: A 5-second window including the moment a mixture of 12.5\% Analyte B is exposed to the array of eight sensors. Machine learning classifiers use discrepancies in sensor resistances responses to discriminate between analytes.}%
    \label{fig:mixup_samples}%
\end{figure*}

\begin{figure*}
    \centering
    \subfloat{{\includegraphics[width=0.40\linewidth]{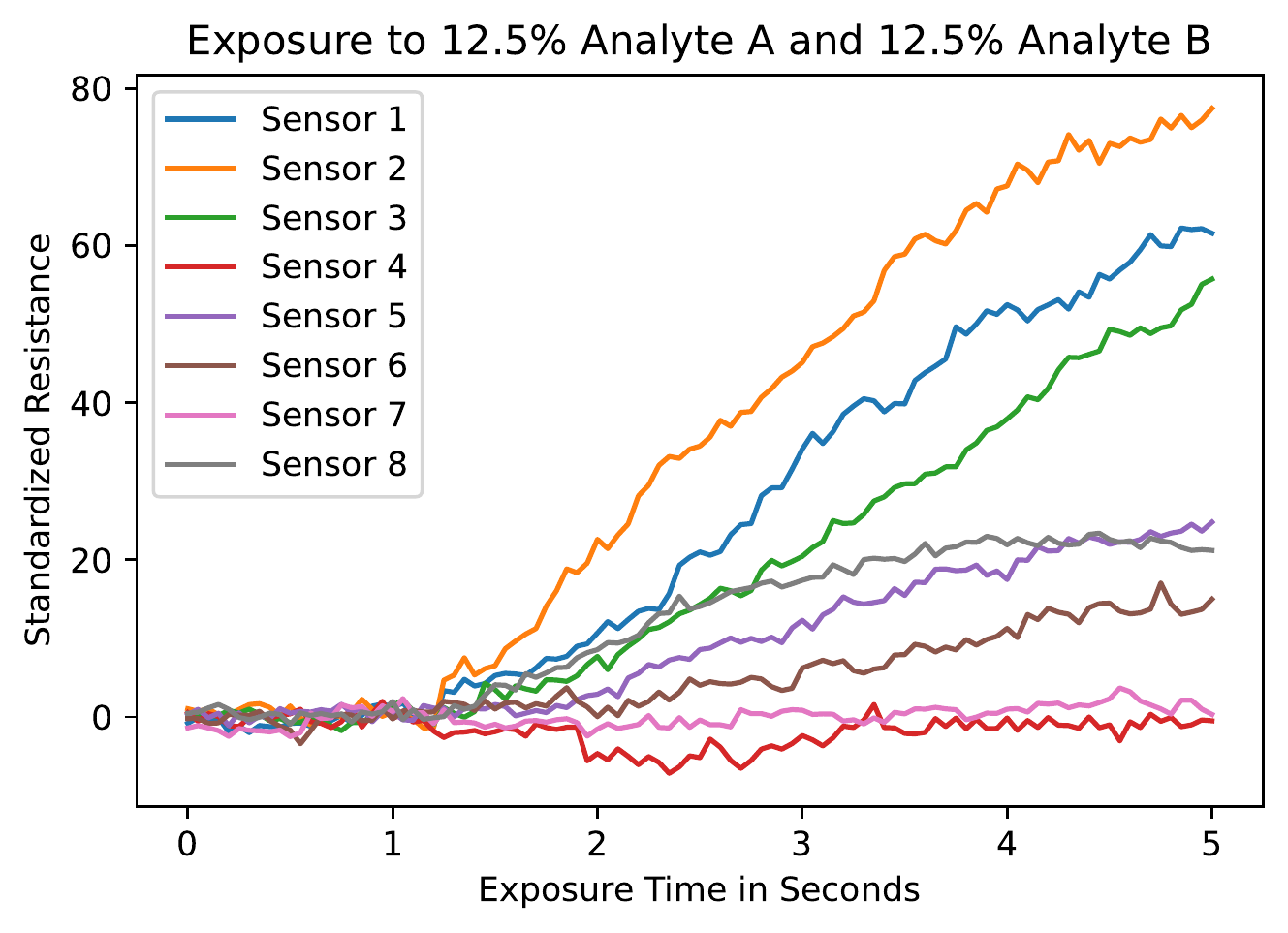} }}%
    \qquad
    \subfloat{{\includegraphics[width=0.40\linewidth]{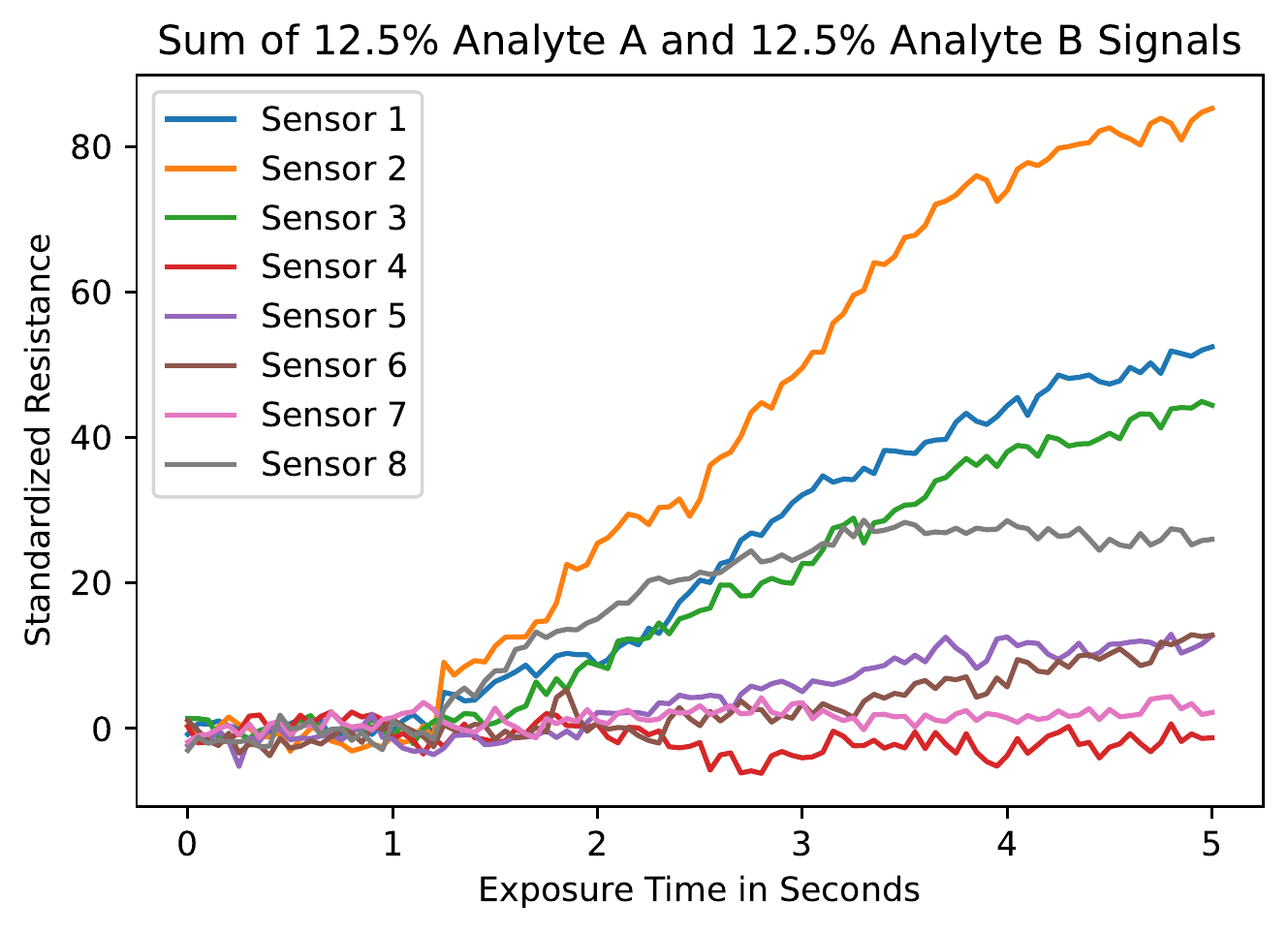} }}%
    \caption{\textbf{Left}: Sensor experiment exposing eight chemiresistive sensors with unique polymer and graphene coatings to a mixture of 12.5\% Analyte A and 12.5\% Analyte B. \textbf{Right}: The summation of 12.5\% Analyte A and Analyte B signals approximates the simultaneous double-analyte exposure to Analyte A and Analyte B. The synthesized double-analyte target in the analyte representation space is taken to be the weighted sum of the two single-analyte representation vectors according to the linear combination weight (Section \ref{sec:lc})}%
    \label{fig:real_lc_doubles}%
\end{figure*}

The data considered here are drawn from two types. Single-analyte exposures are created by exposing one chemical analyte of a given concentration to the multi-array sensor comprising eight polymer and graphene nanoplatelet composites for a defined time period. For this work we trim chemical sensing trials (Figure \ref{fig:annotated}) to exposure windows from 2.4 to 4.0 seconds which includes the moment the chemical analyte first meets the sensor array (Figure \ref{fig:mixup_samples}). The second, more challenging data are the double-analyte exposures: a two-analyte mixture is exposed to the sensor array for a time period. Mixtures containing Analyte A are labelled the positive class for classification. These double-analyte exposures are substantially more difficult to classify as the desired information (the presence of Analyte A) is obscured by an additional analyte which alters the sensor response profiles (Figures \ref{fig:mixup_samples}, \ref{fig:real_lc_doubles}).

In addition to being harder to classify, mixture data are substantially harder to exhaustively gather. The experimental cost of datasets even with limited numbers of analytes and concentrations quickly becomes intractable as trials require expert supervision, experimental design, hardware setup, and data verification processes. Given a number of chemical analytes of interest $x_1,...,x_n$ at various concentrations $c_1,...,c_k$, the number of experiments necessary to perform experimentation of all mixtures at all concentrations is given by 

\begin{equation}\label{eqn:num_experiments_needed}
    \sum_{p=1}^n \binom{n}{p}k^p
\end{equation}

\noindent where $p$ represents the number of analytes in the mixture. Figure \ref{fig:num_experiments} visualizes the complexity of gathering representative data sets for even a limited number of analytes and concentrations: in this case, simultaneous exposure of one, two, three, or four chemical analytes each at one of four potential vapor concentrations. By utilizing single-analyte exposures canonical dimensions of multi-analyte mixture space and incorporating external chemistry knowledge, supervised learning outcomes are improved and constraints to experimental hardware, budget, and time limit may be mitigated.

\section{Relevant Work.}\label{sec:related_work}
Partially due to the meteoric rise of deep learning approaches for natural image and natural language processing, many practitioners are more intensely researching the application of machine and deep learning to applied chemistry which was originally met with skepticism by chemists practicing hand-coded heuristics such as the Morgan Fingerprint \cite{mater2019, morgan1965generation}. Researchers have seen rising application of machine learning to the physical sciences, with ``many demonstrating significant improvements in predictive accuracy and ability to replicate human decision making'' \cite{mater2019, chen2018rise}. For an excellent review of deep learning applications to chemistry and the reception history of chemical applications, please refer to \cite{mater2019} as well as \cite{Schmidt2019} for an excellent survey of surface chemistry. For an introduction to machine learning and its applications for chemists we recommend \cite{mueller2016}.

A primary concern of computation for applied chemistry is the representation of chemical analytes in machine-readable format. This topic has been relevant to chemical researchers for over 160 years - long before even the advent of electronic computation \cite{David2020, wiswesser1968107}. In the deep learning field as in chemistry, representations must serve the purpose of encoding meaningful elements of the object they represent and for machine and deep learning must be processable with vector mathematics. The question of representation poses as many challenges to the encoding of molecules and compounds as it does to images and written text. No singular representation will be appropriate for every task, so countless molecular descriptors, embeddings, and encoding techniques have been proposed.

One such chemical representation is SMILES \cite{Weininger1988}, which encodes molecular structure to ASCII strings given by depth-first traversal over molecular bonds. While the SMILES approach does represent molecules in a machine-readable manner, deep learning researchers have recently relied on the implicit \textit{representation learning} that occurs in multi-layer networks as the source of information density in vector representations \cite{bengio2012}. The feedforward deep neural network outlined in Chemception \cite{goh2017} was trained on a set of molecular images derived from SMILES strings to predict the toxicity, activity, and solvation properties of the input. As a consequence of implicit representation learning in deep neural networks, the penultimate activations of this model may be used to represent molecular analytes in dense real-valued vector embeddings. The Chemception embeddings encode semantically meaningful distances between points which facilitate vector mathematics and downstream learners for fine-tuned tasks (Figure \ref{fig:chemception_embeddings}). Chemception and Mol2vec \cite{jaeger_fulle_turk_2017} use modern deep learning practice to outperform the earlier Morgan Fingerprints representation \cite{morgan1965generation} using pattern recognition on a massive corpus of chemistry data.

Contemporary research utilizes deep learning for chemistry, and in particular compound representations for supervised learning such as drug interaction \cite{ryu2018deep, cai2019deep}, new material discovery \cite{wei2019, Schmidt2019}, and drug design \cite{Yang2019, Gebauer_2022, elton2019, Rigaioglu2020}. These approaches utilize pretrained or novel deep learning embedding models to encode an unseen set of chemical analytes. The encodings are used as information-dense representations of training data for an external task outside of the initial scope of the embedding model. This process is called fine-tuning or transfer learning in the deep learning literature \cite{weiss2016survey, zhuang2020comprehensive}.

Our surveying efforts above report many other techniques utilizing molecular and compound embeddings for downstream supervised learning. However, to our knowledge our ChemVise approach (Section \ref{sec:method}) utilizing transfer learning representations as targets to train embedding models for arbitrary input data remains unexplored in chemical sensing and materials science. Natural image and natural language researchers \cite{frome2013, socher2013, radford2021} have emphasized semantically meaningful target spaces for supervised and generative deep learning. Sentence-to-image generation and image-to-sentence captioning are a promising recent breakthrough which large language transformers have had incredible recent successes \cite{dalle2_2022}. Multi-modal deep representation learners use paired encoder-decoder models which operate in the same meaningful embedding space, and learn to link the two embeddings with a translation model. Rather than natural image and sentence semantics, however, our ChemVise approach utilizes deep representations of molecular SMILES images encodings paired with chemiresistive sensor signals to learn improved latent representations for downstream predictors (Figure \ref{fig:chemception_embeddings}).

\section{Data.}\label{sec:data}
\begin{figure}
    \centering
    \includegraphics[width=0.80\linewidth]{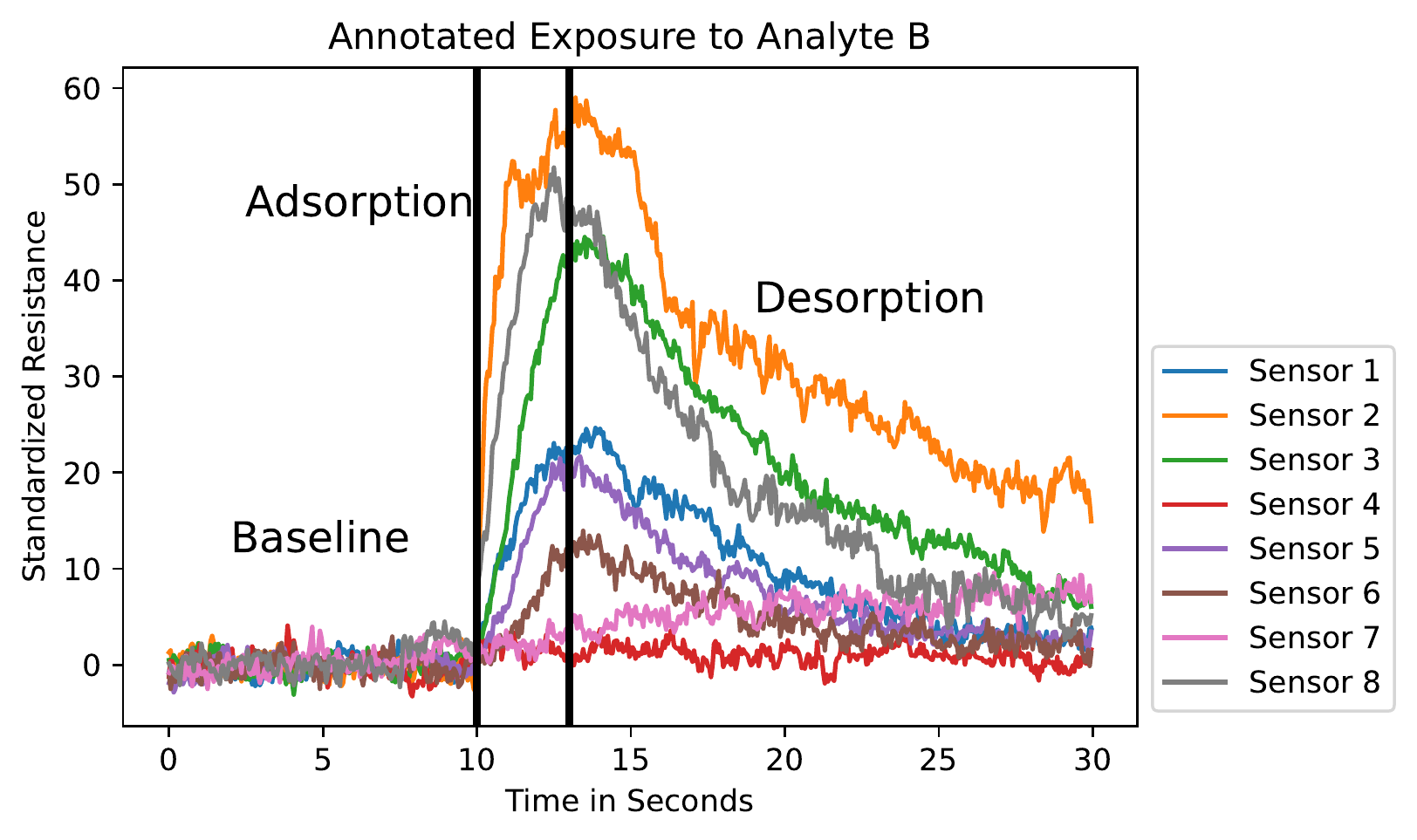}
    \caption{One experimental sample including baseline, absorption, and desorption for an exposure of Analyte B. The sensor resistances rapidly increase during the analyte exposure period as chemical analytes adsorb onto sensor coatings before returning to baseline levels during desorption as analytes diffuse from the surface.}
    \label{fig:annotated}
\end{figure}

Chemical sensors utilize resistance changes of electrodes with unique coatings. Different vapor analytes have different adsorption affinities to the different coatings, resulting in differential resistance changes. Figure \ref{fig:annotated} visualizes the temporal regions of a full analyte exposure experiment. Analyte exposure experiments utilizing chemiresistive sensors contain three exposure periods \cite{nix2022}:

\begin{enumerate}
    \item Controlled flux of analyte and adsorption to the active surface or substrate.
    \item Change in electrical resistance due to interaction of sensor coating with adsorbed and/or absorbed analyte molecules.
    \item Cessation of analyte flux followed by flux of zero gas. Desorption of analyte from sensors which reverses resistance change.
\end{enumerate}

In our research, we utilize an 8-sensor chemiresistive array with chemically diverse polymer and graphene composites to maximize analyte discriminability as in \cite{nallon2016discrimination, wiederoder2017graphene, weiss2018applications}. The data and results here are one set of experiments performed with this array, but the deep learning methodology for transferring analyte representations and supervised classification of multiple analytes may be extended to any chemical sensor array.

Each of the eight sensors utilizes some baseline resistance of electric current through the sensor coating. Different adsorption of analytes leads to the discriminability based on sensor responses seen in the resistances changes between an exposure to Analyte A and Analyte B in Figure \ref{fig:mixup_samples}. These resistance responses are preprocessed from the data collection step with only z-scoring.

In addition to single-analyte exposures of one analyte to the sensor array, multiple analytes may be exposed simultaneously. Here we refer to a two-analyte mixture exposed to the sensor device as a double-analyte exposure, and may be a combination of any two analytes with any relative magnitude. In our binary classification results \ref{sec:results}, a mixture containing any amount of Analyte A is treated as a positive sample and all other mixture pairs are treated as negative samples.

\section{ChemVise Method.}\label{sec:method}
Our proposed ChemVise \footnote{Further detail on the implementation and training of ChemVise along with a PyTorch-style pseudocode is available in the extended materials section.} approach to chemical sensing borrows from the deep learning literature on zero-shot learning \cite{larochelle2008}, transfer learning \cite{weiss2016survey}, and domain transfer \cite{dalle2_2022} to utilize a transformed target space given by an expert model (Figure \ref{fig:chemvise_embeddings}). DeVise \cite{frome2013} used a pretrained skip-gram language embedding model and a pretrained deep convolutional neural network to embed images and their text labels as high-dimensional real-valued vectors in two distinct representation spaces. A translation model is then trained to map between an image-representation space to a label-representation space, thereby linking the spaces and allowing zero-shot semantic prediction of input images.

\begin{figure}
    \centering
    \includegraphics[width=0.95\linewidth]{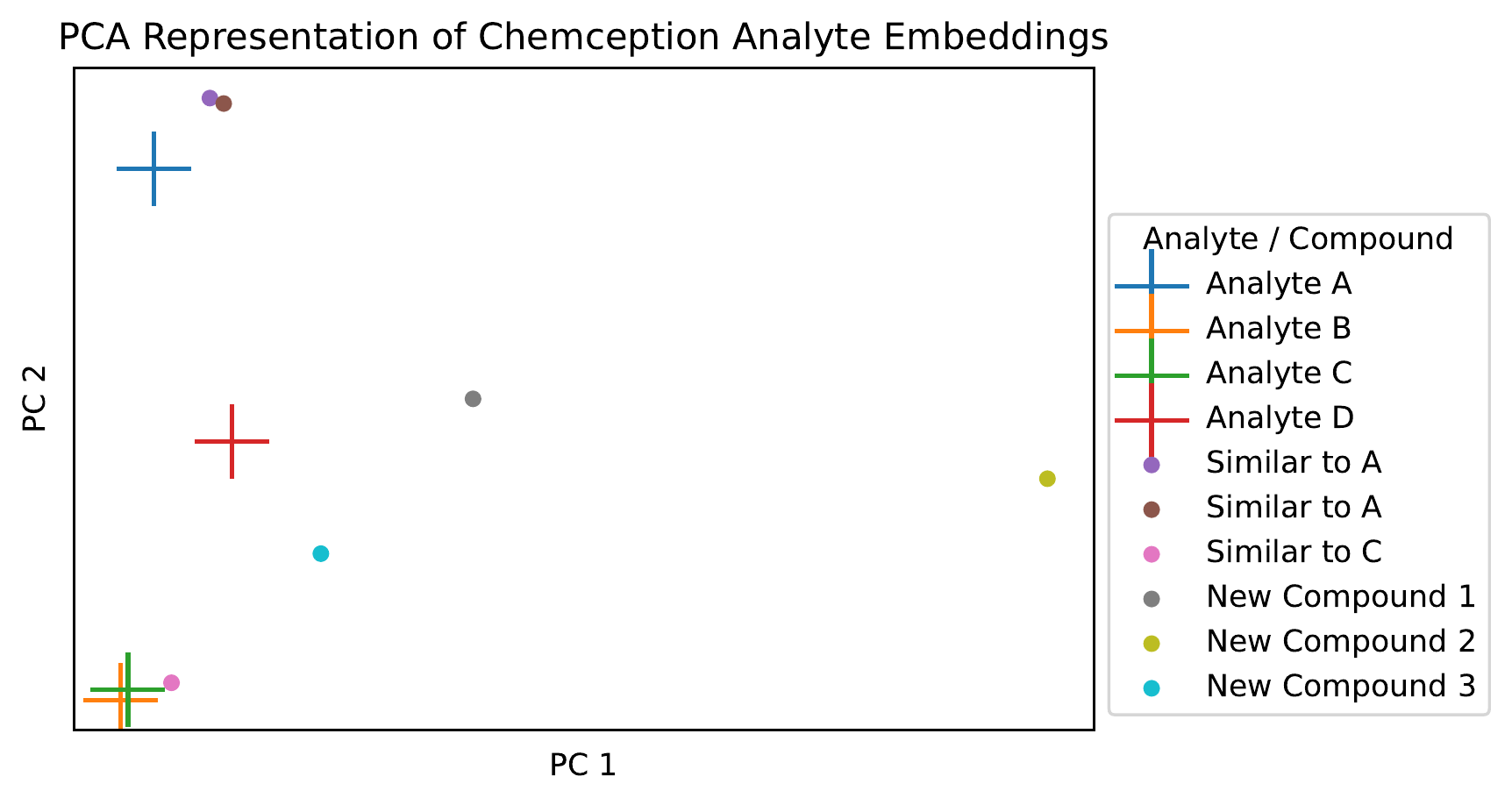}
    \caption{Embeddings of four analytes of interest as well as six comparison analytes and compounds. The penultimate activations of the Chemception model are used as embeddings and reduced to two dimensions with PCA for inspection. Chemception provides representations of analytes used as targets for the ChemVise embedder.}%
    \label{fig:chemception_embeddings}%
\end{figure}

There exist many pretrained deep representation models trained on large chemistry corpi mentioned in Section \ref{sec:related_work}. In this work we utilize penultimate activations of the pretrained Chemception \cite{goh2017} supervised model which was trained to predict toxicity, activity, and solvation of a large dataset of chemicals. The inherent representation learning of deep neural networks provides representations of analytes we use to create a more meaningful target space for the ChemVise embedding model which trained from scratch for our novel chemical sensing hardware.

ChemVise uses targets represented by embeddings rather than traditional labels for a chemiresistive sensor signal input. Using representations of analytes as a target rather than typical one-hot classification or multiple regression targets leads to improved decision boundaries and signal representations and can be adapted to arbitrary sensing hardware. Figure \ref{fig:chemception_embeddings} shows 2D visualizations of embeddings of analytes given by Chemception \cite{goh2017} encodings of analytes represented as images of SMILES strings match our intuition about similarity of analytes in a high-dimensional space. Representation spaces replace the label space as the target for a high-dimensional output.

\begin{figure*}
    \centering
    \subfloat{{\includegraphics[width=0.45\linewidth]{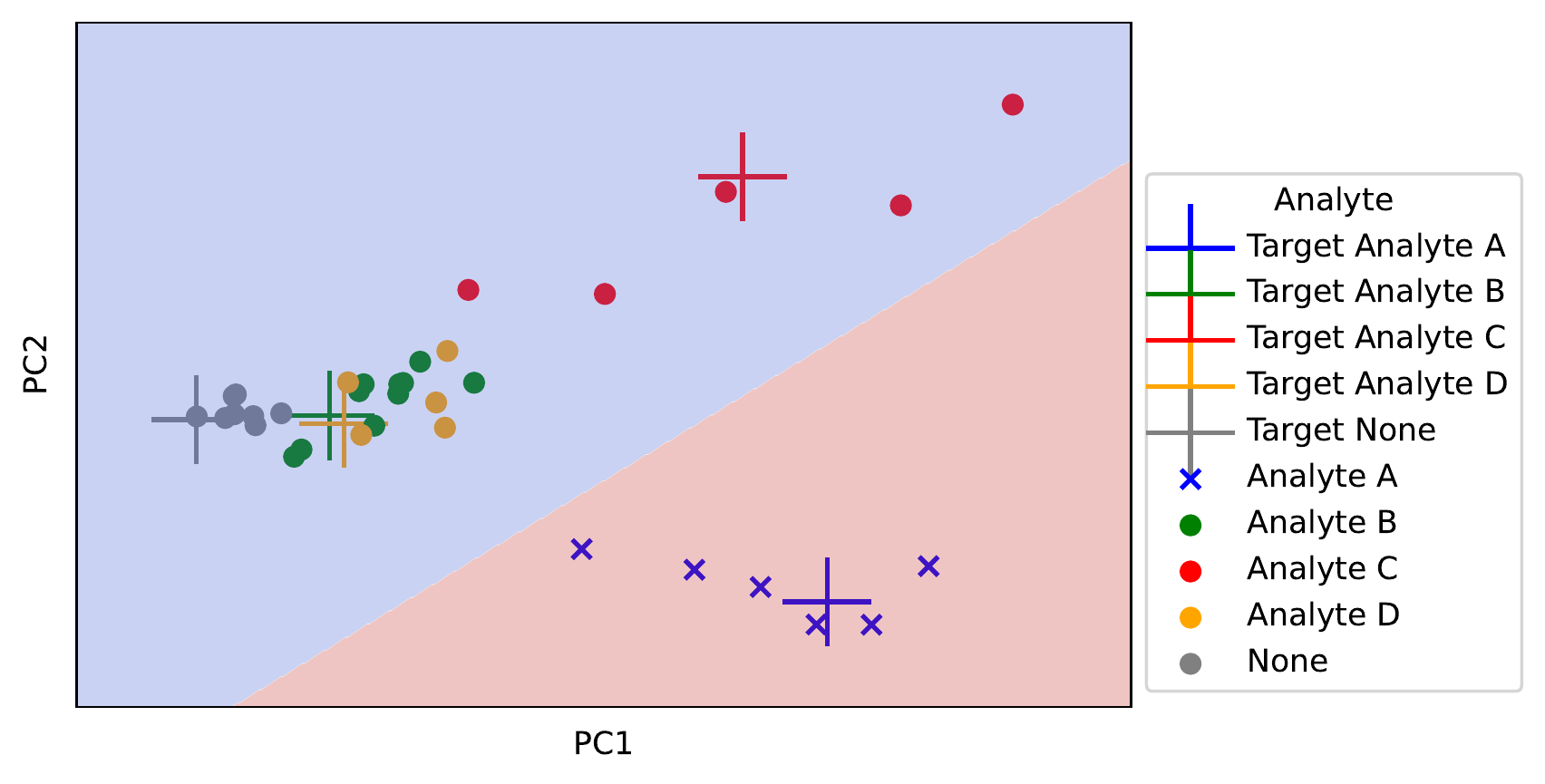} }}%
    \qquad
    \subfloat{{\includegraphics[width=0.45\linewidth]{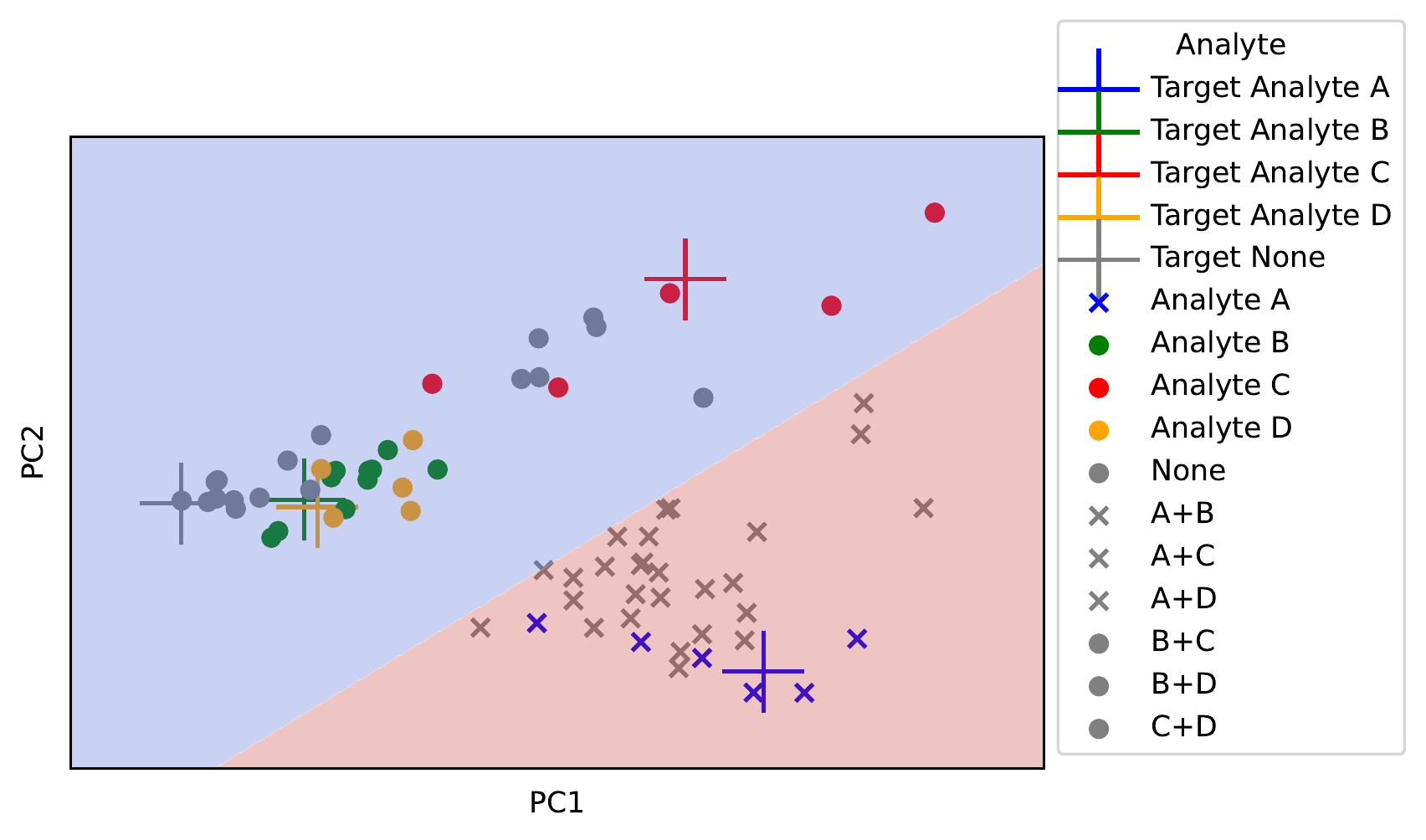} }}%
    \caption{This visualization shows the validation singles and testing doubles (black) embedded by the ChemVise model then classified with a linear SVC. This image is for visual illustration only as ChemVise utilizes a 512-dimensional representation space, which is more accurate than the same model under a 2-dimensional PCA decomposition. \textbf{Left}: ChemVise embeddings given training the deep learning model to map signal samples to their corresponding target given by the matching color ``+''. The tunable decision boundary linearly separates the positive and negative samples. \textbf{Right}: Double-analyte samples under this trained embedding model fall in line with the correct classification given by the SVC trained with only single-analytes. This embedding outperforms simple decomposition embeddings by incorporating molecular semantics into the embedded representations, and allows even double-analyte samples to be linearly separated.}%
    \label{fig:chemvise_singles-doubles_embeddings}%
\end{figure*}

A deep fully-connected neural network is trained to map from the input data space of sensor signals to this molecular-semantics representation space. Any supervised learning algorithm with a multidimensional real-valued output may be used. The ChemVise process provides a trained embedder which encodes any sensor response input in the information-dense space with improved separability between classes. Subsequently any classifier can be used in the chemistry space for improved classification outcomes (Figure \ref{fig:chemvise_singles-doubles_embeddings}) compared to non-embedded samples (Figure \ref{fig:pca_svc}).

Each point of an analyte representation space encodes a similarity and dissimilarity to other molecule and compound points. ChemVise must provide a faithful embedding to the properties of chemical compounds as well as multi-analyte mixture signals. For this reason we implement linear interpolation between single-analyte exposures with synthetic targets given by the geodesic center of the two canonical elements.

PyTorch-style pseudocode for the implementation of ChemVise is provided. Since the implementation of the sensor hardware and capture devices may change, this algorithm paired with any pretrained chemical representation model for analyte targets is compatible with arbitrary input signals.

\section{Linear Combinations of Chemiresistive Signals.}\label{sec:lc}

We utilize linear combinations of single-analyte exposures as approximations to multi-analyte signals which are harder to gather. Linear combinations are sometimes utilized in natural image data augmentation designed to promote simple linear behavior in-between training examples for deep neural networks \cite{mixup2017}. The data-agnostic linear combination augmentation synthesizes samples given by linear interpolations between random examples and their training labels:
$$
  \Bar{x} = \lambda x_i + (1-\lambda)x_j  \\
  \Bar{y} = \lambda y_i + (1-\lambda)y_j
$$

\noindent Where $y_i, y_j$ are analyte exposure embedding vectors corresponding to sensor response vectors $x_i, x_j$.

Linear combinations as applied to chemical sensing approximates real double-analyte chemical sensor exposures using only single-analyte exposures as training data. We modify the typical probability density function from a Beta distribution for ImageNet optimization \cite{mixup2017, krizhevsky2012imagenet} to a tunable uniform distribution between $a$ and $b$. The maximum and minimum of the distribution may be tuned to account for what level of mixture sensitivity should be classified as a positive sample. Here, we utilize $(a,b) = (0.3, 0.7)$ as the bounds for mixtures. Otherwise, utilization remains the same as in the natural image paradigm wherein samples are linearly interpolated during training using linear combinations of training pairs.

\section{Results.}\label{sec:results}
We provide results for a training paradigm given by training on single-analyte exposures and testing on challenging out-of-distribution obscured double-analyte exposures. We demonstrate the importance of representation selection (Section \ref{sec:representation_matters}) as well as optimized models for the rapid detection of obscured Analyte A given variable-length exposure windows (Section \ref{sec:exposure_results}).
Results within follow a consistent algorithm (K-fold validation hyperparameter gridsearch) for the selection of hyperparameters prior to observation of the holdout testing data using the following from \cite{moore2022cycles}:
\begin{enumerate}
    \item ChemVise
    \begin{itemize}
        \item Model width: 128, 512, 1024, 2048, 4096, 8192
        \item Learning rate: 1e-7, 5e-7, 1e-6, 5e-6, 1e-5
        \item Training epochs: 1000, 2000, 4000
        \item Batch size: 4, 8, 16, 32
    \end{itemize}
    \item Feedforward Neural Networks \cite{rumelhart1986learning}
    \begin{itemize}
        \item Model width: 128, 512, 1024, 2048, 4096, 8192
        \item Learning rate: 1e-7, 5e-7, 1e-6, 5e-6, 1e-5
        \item Training epochs: 1000, 2000, 4000
        \item Batch size: 4, 8, 16, 32
    \end{itemize}
    \item Baseline models:
    \begin{enumerate}
        \item Support Vector Machine \cite{cortes1995support}
        \begin{itemize}
            \item ``C'' Penalty: uniform(1e-4, 8e-3)
            \item Kernel: ``Linear''
            \item Kernel Degree: random(0, 8)
            \item Class Weight (Negative:Positive): 1:1, 1:2, 1:4
        \end{itemize}
        \item XGBoost\footnote{Tree boosting} \cite{Chen_2016}
        \begin{itemize}
            \item Column Ratio: random(1,8)
            \item Gamma: uniform(0,0.5)
            \item Learning Rate: uniform(0.03, 0.3)
            \item Max Depth: random(1,8)
            \item N Estimators: random(50,800)
        \end{itemize}
    \end{enumerate}
\end{enumerate}

\subsection{Representation Matters.}\label{sec:representation_matters}

\begin{figure}
    \centering
    \includegraphics[width = 0.9\linewidth]{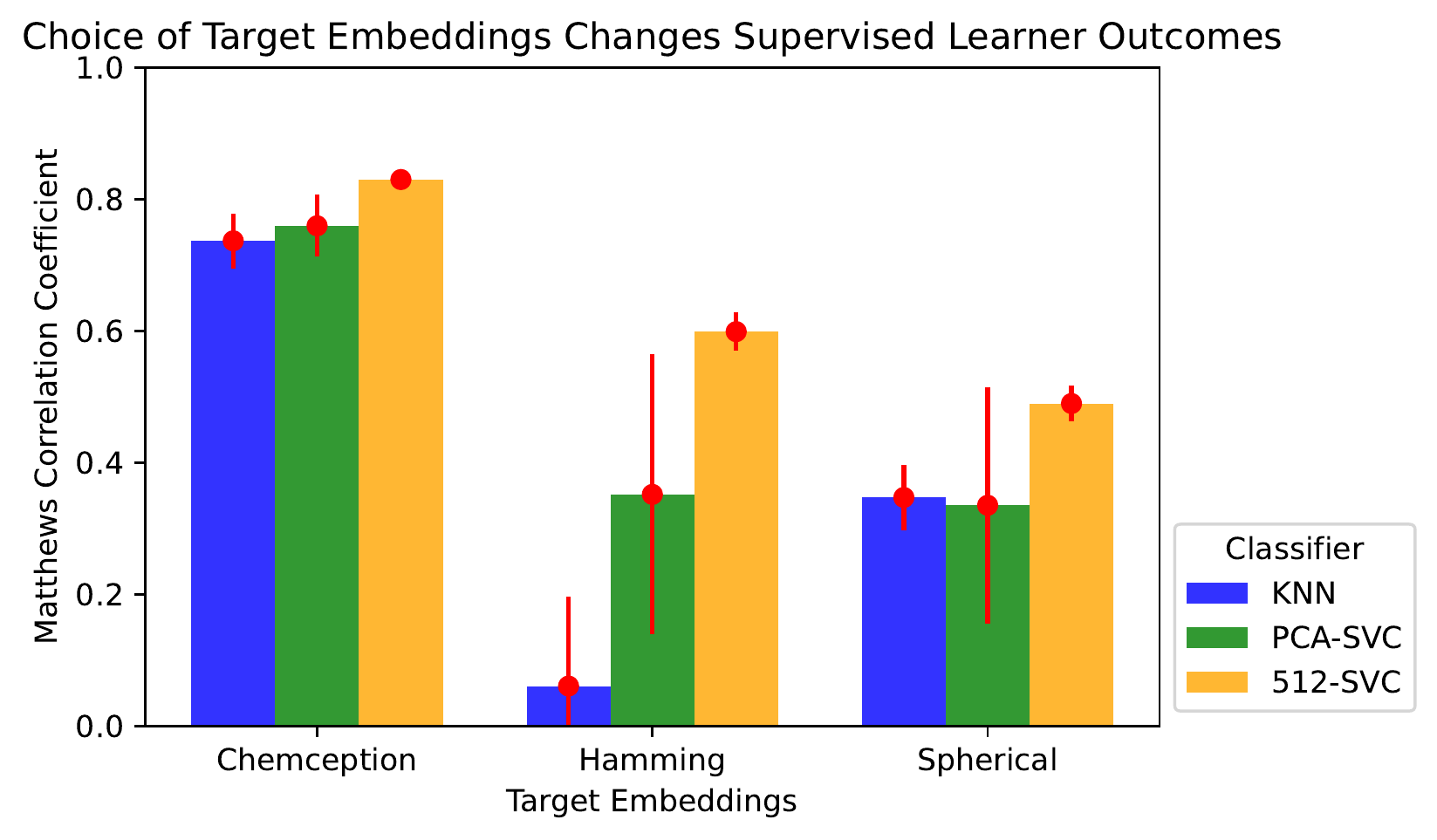}
    \caption{Classification metric on double-analyte exposures as a function of embedding model and downstream classifier. Downstream classifiers include a K-Neighbors classifier in the representation space, or a Support Vector Classifier in the representation or 2-dimensional projection of the representation space. The choice of Chemception, one-hot, or equidistant spherical targets modifies the training of the embedding deep learning model prior to classification by the classifier head.}
    \label{fig:choice_of_target_embedding}
\end{figure}

Though our design utilizes the pretrained Chemception \cite{goh2017} to represent analyte SMILES images as meaningful vectors in a target space, in theory any space could be used to represent analyte labels for embedder training. Here we compare the Chemception representation space with two alternative analyte representation spaces to use as targets for the embedding prior to SVC classification. ``One-hot'' denotes utilizing 512-dimensional (matching the Chemception embedding dimension) representations for analytes where signals containing Analyte $i$ are represented as $e_i$, where $e_i$ is the zero-vector with $1$ at index $i$. ``Equidistant spherical'' denotes utilizing equidistant points from the 512-dimensional unit hypersphere as representations for the chemical analytes.

Figure \ref{fig:choice_of_target_embedding} demonstrates the importance of the semantically meaningful space for the ChemVise embedding. Whereas one-hot and equidistant spherical representation spaces struggle to provide meaningful representation spaces for downstream SVC classification, Chemception consistently provides a high-quality representation space for three candidate classifiers. Semantic meaning implemented as intra-analyte distances dramatically improves the decision boundary outcomes over two approaches which do not incorporate external chemistry knowledge.

\subsection{Variable Exposure Times.}\label{sec:exposure_results}
\begin{figure}
    \centering
    \includegraphics[width=0.90\linewidth]{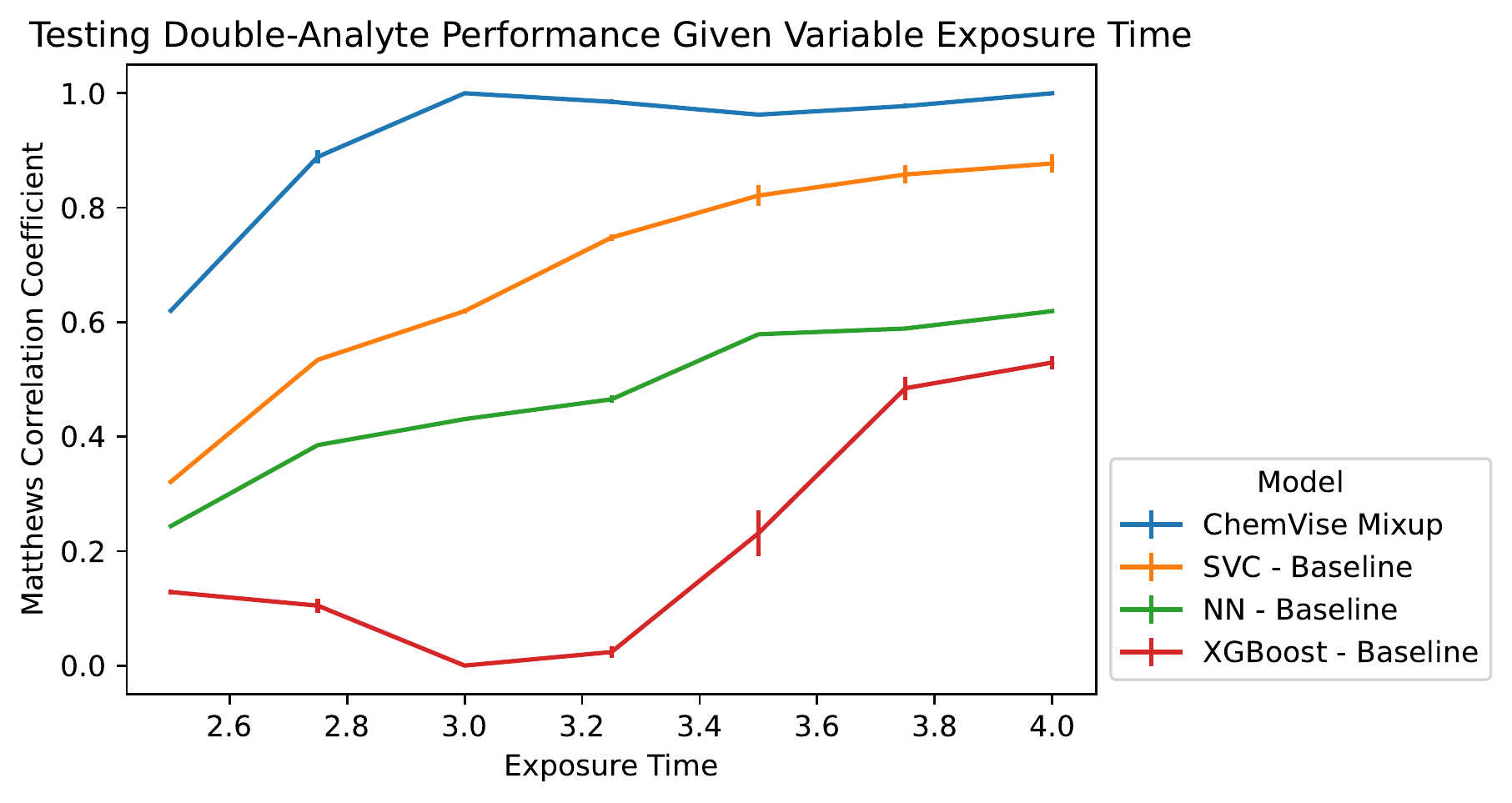}
    \caption{Matthews Correlation Coefficient score given the prediction of Analyte A in a holdout set of 39 real double-analyte experiments after training on only single-analyte exposures. Given five hyperparameter optimizations over five 5-fold cross validation experiments, model performances for 7 different exposure window lengths from 2.4 to 4.0 seconds.}
    \label{fig:results_sec}
\end{figure}

In Figure \ref{fig:results_sec} we demonstrate results on the rapid classification of Analyte A in the presence of obscurant analytes with no corresponding obscured-analyte training data. Though multiple machine and deep learning approaches show a positive scaling behavior in the response metric as a function of exposure time, the ChemVise approach substantially and consistently outperforms the alternative models for the out-of-distribution detection task.

\section{Conclusion.}\label{sec:conclusion}

ChemVise representation learning outperformed linear decomposition embeddings as well as standard machine and deep learning supervised models by incorporating molecular semantics into the embedded representations. We demonstrated how improved chemical representations incorporating domain knowledge on molecular semantics for chemical sensing allow double-analyte samples to be encoded such that they are linearly separable with simple classifiers when only single-analyte real experiments are available. This approach may benefit researchers performing chemical detection with arbitrary hardware designs, and does not require pretrained models in the input domain. Rather, chemical classification may utilize sensor data from any sensors provided a corresponding target domain exists and is sufficiently studied.

Though transfer learning may be appropriate when the input data lies in a well-studied domain with learnable representations, relying on both transfer learning and ChemVise-style target adaptation dramatically increases the number of tasks which may incorporate exterior data corpi. The ChemVise approach may extend the toolbox of a machine learning engineer performing supervised learning on a more diverse set of tasks with diverse inputs. The implementation would retain the deep embedding model and replace the target space with an appropriate pretrained model capable of representing the targets of the data set with some meaningful distance encoding. The generalizable approaches to supervised learning discussed here may dramatically improve sensor machine learning results and move away from complex, hand-tuned representations of domain knowledge.

\bibliographystyle{abbrv}
\bibliography{chemvisebib} 

\end{document}